\documentclass{article}

\usepackage[utf8]{inputenc} 
\usepackage[T1]{fontenc}    
\usepackage{hyperref}       
\usepackage{url}            
\usepackage{booktabs}       
\usepackage{amsfonts}       
\usepackage{nicefrac}       
\usepackage{microtype}      
\usepackage{xcolor}         
\usepackage{amsmath,amssymb, amsthm}
\usepackage{enumitem}
\usepackage{algorithm}
\usepackage{algpseudocode}
\usepackage{environ}
\usepackage{array}
\usepackage{graphicx}
\usepackage{caption}
\usepackage{subcaption}
\usepackage{physics}
\usepackage{geometry}
\usepackage{mathtools}
\usepackage{thmtools}
\DeclareMathOperator{\argmax}{arg\,max}
\allowdisplaybreaks
\numberwithin{equation}{section}
\newtheorem{theorem}{Theorem}
\newtheorem{lemma}{Lemma}
\declaretheorem[name=Lemma,numbered=no]{lemmarestate}
\declaretheorem[name=Theorem,numbered=no]{theoremrestate}

\title{Multi-Agent Best Arm Identification in Stochastic Linear Bandits}

\author{
  Sanjana Agrawal\\
  Department of Computer Science\\
  Indiana University Bloomington\\
  \texttt{sanagra@iu.edu} \\
  \and
  Sa\'ul A. Blanco \\
  Department of Computer Science\\
  Indiana University Bloomington\\
  \texttt{sblacor@indiana.edu} \\
}

\date{May 24, 2025}

\begin{document}

\maketitle

\begin{abstract}
  We study the problem of collaborative best-arm identification in stochastic linear bandits under a fixed-budget scenario. In our learning model, we first consider multiple agents connected through a star network, interacting with a linear bandit instance in parallel. We then extend our analysis to arbitrary network topologies. The objective of the agents is to collaboratively identify the best arm of the given bandit instance with the help of a central server while minimizing the probability of error in best arm estimation. To this end, we propose two algorithms, \textsc{MaLinBAI-Star} and \textsc{MaLinBAI-Gen} for star networks and networks with arbitrary structure, respectively. Both algorithms utilize the technique of G-optimal design along with the successive elimination based strategy where agents share their knowledge through a central server at each communication round. We demonstrate, both theoretically and empirically, that our algorithms achieve exponentially decaying probability of error in the allocated time budget. Furthermore, experimental results on both synthetic and real-world data validate the effectiveness of our algorithms over the state-of-the art existing multi-agent algorithms.
\end{abstract}

\section{Introduction} The multi-armed bandit problem (MAB) is a classic framework in sequential decision making, capturing the exploration-exploitation trade off faced in many real-world domains, for example, recommendation systems \cite{ gentile2014online, li2010contextual,li2016collaborative}, clinical trials \cite{durand2018contextual, wang1991sequential}, online advertising \cite{tao2018best}, adaptive routing \cite{awerbuch2008online} and so on. An instance of MAB problem consists of a set of possible choices called arms. The learning agent sequentially chooses an arm and receives a reward related to the chosen arm. The goal of the agent is to either maximize the cumulative reward (equivalently, minimize the regret) over the time, referred as \textit{regret minimization} problem \cite{bubeck2012regret,cesa2013gang,lattimore2020bandit} or, to identify the best arm within a specified constraint. The latter variant is known as the \textit{best-arm identification} or \textit{pure exploration} problem, which is studied in two different settings based on the specific constraint: (1) \textit{fixed-budget} \cite{audibert2010best,bubeck2009pure, karnin2013almost} and (2) \textit{fixed-confidence} \cite{chen2017towards,garivier2016optimal, mannor2004sample}. While the fixed-budget setting aims to identify the best arm with smallest error probability within the allocated time budget, the goal in fixed-confidence setup is to identify the best arm with the given confidence level using minimum arm pulls. \\

\noindent In this paper, we study fixed-budget best-arm identification in \textit{stochastic linear bandit} (SLB) \cite{abbasi2011improved,auer2002using}. In SLB setup, the arm set is a subset of $\mathbb{R}^d$ and pulling an arm yields a reward which is a noisy observation of the linear combination of pulled arm and an unknown vector $\theta \in \mathbb{R}^d$. The best arm is defined as the arm vector closest to vector $\theta$ in $\mathbb{R}^d$ space. Various real-world scenarios can be effectively represented using the framework of stochastic linear bandits. For example, in online advertising \cite{tao2018best}, where each advertisement option can be modeled as an arm representing the properties like genre, keywords, duration and so on. The common vector $\theta$ can encapsulate the characteristics of the targeted audience such as age group, geographic locations, etc.\\

\noindent In many such applications, it is appealing to use collaboration and communication in order to speed up the learning process. For instance, in web recommendation applications, increased amount of data and high volume of user requests overwhelms a single processor's capacity to handle learning tasks efficiently. In such scenarios, it is desirable to deploy multiple servers such that each user request is routed to one of the servers. These servers can then collaborate by sharing their local insights to collectively improve the recommendation process. Such applications have motivated the development of multi-agent collaborative algorithms where multiple autonomous agents work together to solve a common learning task (see, for example, \cite{do2023multi, ghosh2022multi,he2022simple, SW08, XJHTZ22}). Recently, \cite{wang2023pure} developed a \emph{federated} algorithm for pure exploration in linear bandits under fixed-confidence framework. However, the problem of collaborative best-arm identification in linear bandits under fixed-budget setting still remains unaddressed. \\

\noindent In this paper, we address this open problem by introducing two novel algorithms \textsc{MaLinBAI-Star} and \textsc{MaLinBAI-Gen} for linear bandits. Our framework involves a communication model where a group of agents is interconnected through a communication network, collaboratively learn the best arm facilitated by a central server. First, we devise the algorithm \textsc{MaLinBAI-Star} designed for star networks, where $M$ independent agents perform the learning task, exchanging their observations through a centralized server. However, real-world applications often involve agents that are geographically dispersed, making it impractical to establish a central server at a location that can facilitate efficient communication for all agents simultaneously. Consider, for instance, an e-commerce platform selling multiple brands (arms) and trying to determine the best-performing brand for a product \cite{shahrampour2017multi}. The platform's users (agents) may be scattered across different regions, so deploying a single server to serve all users is inefficient. To address this, the company can deploy multiple servers across various locations, with each server dedicated to collecting feedback from users within a specific geographic area, and later, the company aggregates this feedback to make a decision about the best brand. Motivated by such applications, we devise algorithm \textsc{MaLinBAI-Gen} which extends the applicability of \textsc{MaLinBAI-Star} to more complex generic networks. Our algorithm \textsc{MaLinBAI-Gen} takes into account the structure of the communication network using \textit{dominating sets}~\cite{haynes2013fundamentals}.\\


\noindent Given the learning network comprising multiple interconnected agents collaborating to complete a learning task, \textsc{MaLinBAI-Gen} first partitions the network into dominating sets. Each block of the dominating set partition executes \textsc{MaLinBAI-Star} independently where the corresponding dominant node plays the role of central server for that block in the partition. Later, the information gathered at these dominant nodes is aggregated by an ensemble using the majority vote to decide the best-arm of the given bandit instance.\\

\noindent By incorporating the network structure in the learning process, \textsc{MaLinBAI-Gen} manages to achieve same performance level as of \textsc{MaLinBAI-Star} while reducing the communication cost significantly. \\

\noindent Notably, finding the minimum dominating set of a graph is computationally hard. However, there is a well-established body of research offering approximation algorithms for computing the minimum dominating set, as discussed in \cite{haynes2013fundamentals, kuhn2003constant, wieland2001domination}. Nevertheless, the upper bound for the error probability of our algorithm \textsc{MaLinBAI-Gen}, remains independent of the size of the dominating set. In fact, \textsc{MaLinBAI-Gen} is designed to function efficiently with any valid dominating set partition, making it adaptable across a range of network structures.\\

\noindent\textbf{Related work:} Fixed-budget best arm identification in linear bandits has been studied previously for single agent setting \cite{alieva2021robust,hoffman2014correlation, katz2020empirical}. \texttt{BayesGap} \cite{hoffman2014correlation} uses a Bayesian approach which incorporates the correlation among the arms using a Gaussian process. \texttt{Peace} \cite{katz2020empirical} uses a experimental design based on the Gaussian-width of the underlying arm set. Recently, \cite{azizi2021fixed} proposed the first BAI algorithm for generalized linear models. In \cite{yang2022minimax}, authors developed a minimax optimal algorithm OD-LinBAI which uses the G-optimal design for arm selection. A related work in \cite{wu2016contextual} uses a adjacency graph to capture the dependency between users where observed rewards on each user are determined by a group of neighboring users in the graph. However, none of these algorithms incorporate collaborative learning. Recently, \cite{wang2023pure} proposed a federated algorithm to tackle the pure-exploration problem in linear bandits within a fixed-confidence framework, while \cite{du2021collaborative} studied the collaborative pure exploration in kernel bandit model.  Nevertheless, there remains a significant gap in the literature of linear bandits regarding federated pure exploration under a fixed-budget setting. In this paper, we fill this gap by introducing two collaborative algorithms \textsc{MaLinBAI-Star} and \textsc{MaLinBAI-Gen}, designed for fixed-budget best arm identification.\\

\noindent Another area of research focuses on collaborative regret minimization \cite{karpov2024communication,kolla2018collaborative, liu2010distributed, szorenyi2013gossip, zhu2023distributed}. A significant body of work has been devoted to collaborative regret minimization specifically for stochastic linear bandits (SLBs). Most closely related work on federated learning in linear bandits include \cite{amani2023distributed,chawla2022multiagent,do2023multi, dubey2020differentially,ghosh2022multi, he2022simple, huang2021federated, li2022asynchronous,wang2019distributed} addressing the problem of collaborative regret minimization. In particular, \cite{he2022simple,li2022asynchronous} considers the centralized model with asynchronous communication between the server and agents. Moreover, \cite{do2023multi, ghosh2022multi} study the problem under heterogeneous model where each participating agents has its own reward parameter $\theta$. Furthermore, \cite{moradipari2022collaborative} studies the heterogeneous setup in a decentralized setting where agents can communicate with their immediate neighbors. The work in \cite{dubey2020differentially} establishes a decentralized learning algorithm with deferential privacy. Recently, \cite{amani2023distributed} established a lower bound on the communication cost of distributed contextual linear bandit and proposed a minimax optimal algorithm for regret minimization. However, the primary focus of these studies is regret minimization, which does not directly tackle the specific pure exploration problem that we aim to address.\\

\noindent The communication model used for generic networks in our algorithm is close to \cite{kolla2018collaborative} where the authors used the concept of dominating set partition to incorporate the network structure. While \cite{kolla2018collaborative} deal with the problem of collaborative regret minimization, recently \cite{jha2022collaborative} extended their work to multi-agent best arm identification under fixed budget. However, both of those papers focus on classic multi-armed bandits and can not be directly applied to linear bandits. With this paper, we add to this literature and propose the collaborative best-arm identification algorithm for linear bandits.

\noindent\textbf{Main Contributions:} Our main contributions are the following:
\begin{itemize}
    \item We devise an algorithm \textsc{MaLinBAI-Star} for fixed-budget best arm identification in star networks. Building on this, we generalize the approach to arbitrary network topologies, introducing the algorithm \textsc{MaLinBAI-Gen}. This algorithm doesn't make any assumption on the network structure except that the network should remain consistent over the execution time. \textsc{MaLinBAI-Gen} integrates the network's structure into the learning process through dominating set partitions, delivering error bounds similar to those achieved by \textsc{MaLinBAI-Star} while incurring lower communication costs. Further, we show that the upper bound for error probability of our algorithms is near-optimal with respect to the lower bound presented in \cite{yang2022minimax}.

    \item Finally, we conduct extensive numerical experiments on both synthetic and real-world data and compare our algorithms to several baselines including (1) minimax optimal single-agent fixed-budget pure-exploration algorithm OD-LinBAI \cite{yang2022minimax}, (2) multi-agent regret minimization algorithms FedLinUCB \cite{he2022simple}, Async-LinUCB \cite{li2022asynchronous} and, (3) multi-agent fixed-confidence best-arm identification algorithm FALinPE \cite{wang2023pure}. The numerical results demonstrate that our approach stands out notably in contrast to the aforementioned algorithms. 
\end{itemize}

\noindent \textbf{Notation:} If $n$ is a positive integer, $[n]$ denotes the set $\{1,2,\ldots,n\}$. Furthermore, if $x\in\mathbb{R}^d$, then $\|x\|_2$ denotes the 2-norm of $x$ and $\|x\|_A$ denotes $\sqrt{x^{\top} Ax}$. Finally, if $C,D$ are sets, $|C|$ denotes the cardinality of $C$ and $C\setminus D$ denotes their set difference. 


\section{Problem Formulation}\label{sec:problem}
We consider an instance of linear bandits with a finite action set $\mathcal{A} \subseteq \mathbb{R}^d$ with $K$ arms. There are $M$ agents connected through a network and, a central server facilitating the communication between the agents. At each time instance $t$, all agents are given the decision set $\mathcal{A}$. Each agent $m \in [M]$, selects an arm $a_{m,t} \in \mathcal{A}$, and receives a random reward $r_{m,t} = \left<\theta, a_{m,t}\right> + \eta_{m,t}$, where $\theta \in \mathbb{R}^d$ is a fixed but unknown parameter to be learned and, $\eta_{m,t}$ is zero-mean, independent $R$-sub-Gaussian noise with $R \ge 0$. 
Please note that these are the standard assumptions in linear bandit literature \cite{abbasi2011improved}.\\

\noindent For any arm $a$, its expected reward is given by $\left<\theta, a\right>$. For simplicity, we assume that there is a unique best arm $a^*$ with highest expected reward. For any other suboptimal arm $a_i$, $\Delta_i = \left<\theta, a^* - a_i\right>$ represents its \emph{sub-optimality gap}, and $\Delta_{\min}$ denotes the minimum sub-optimality gap of the given instance, i.e., $\Delta_{\min} = \min_{a_i \in \mathcal{A}}\Delta_i$. The goal of the agents is to collaboratively learn the arm with the highest expected reward, namely $a^* = \argmax_{a\in \mathcal{A}} \left<\theta, a\right>$, in the given time budget $T$. 

\subsection{Preliminaries} 
In order to determine the best arm, the algorithm needs to sequentially estimate the unknown vector $\theta$ using the observation history of the agents. In our algorithms, we use the ordinary least-squares estimator to estimate $\theta$. Let $a_{1}, a_{2}, \ldots, a_{t}$ and $r_{1}, r_{2}, \ldots, r_{t}$ be the information at the server sent by the agents regarding the arms played and corresponding rewards obtained till time $t \in [T]$, the estimate of $\theta$ is given by
\begin{equation}\label{theta_estimate}
    \hat{\theta}_{S} = V_S^{-1}D_S,
\end{equation}
where
\begin{equation}\label{local_hist}
    V_{S} = \sum_{s=1}^t a_{s}a_{s}^{\top}, \text{ and }  D_{S} = \sum_{s=1}^t a_{s}r_{s}.
\end{equation}

\noindent Using this estimator of $\theta$, the variance of the estimated expected reward of an arm $a \in \mathcal{A}$ is given by the matrix norm $\|a\|_{V_S^{-1}}$. In order to minimize the probability of error in best-arm identification, the algorithm needs to minimize the maximum value of this matrix norm across all the arms. Note that it is equivalent to finding a G-optimal design for $\max_{a \in \mathcal{A}}\|a\|_{V_S^{-1}}$ with respect to $V_S^{-1}$.\\

\noindent Formally, as defined in \cite{pukelsheim2006optimal}, the problem of finding a G-optimal design seeks to find a probability distribution $\pi$ on the action set satisfying $\pi: \mathcal{A} \rightarrow [0, 1]$ and $\sum_{a \in \mathcal{A}} \pi(a) = 1$ that minimizes:
\begin{equation}\label{g_opt}
   g(\pi) = \max_{a \in \mathcal{A}}\|a\|_{V_{(\pi)}^{-1}}^2 \text{, where }  V_{(\pi)} = \sum_{a \in \mathcal{A}} \pi(a)aa^{\top}.
\end{equation}
It is worth noticing that finding an exact solution to the G-optimal design problem is computationally hard. However, in our case, it suffices to have an approximate solution to the problem. In our algorithms, we compute the 1-approximate G-optimal design using the Frank-Wolfe based algorithm given in \cite{todd2016minimum}.\\

\noindent Finally, for the OLS estimate to be well defined, the matrix $V_S$ needs to be nonsingular. This requires that the active set of arms span the entire space $\mathbb{R}^d$. However, in our phased elimination strategy, the active set of arms may not span $\mathbb{R}^d$ in the later stages of the algorithm. To resolve this issue, we project the arms onto a lower dimensional space such that the matrix $V_S$ becomes non-singular.

\section{Algorithm for Star Networks}\label{section:star}
In this section, we describe our algorithm \textsc{MaLinBAI-Star} for star networks. Our learning model consists of a star network with a central server and $M$ agents. Each agent has a direct link to the central server. However, the setup assumes no direct links between the agents themselves. In other words, every agent in the network can directly communicate with the server but no two agents can communicate with each other. \\

\noindent The pseudocode of \textsc{MaLinBAI-Star} is presented in Algorithm~\ref{alg:star}. Given the time budget $T$, our algorithm partitions the budget into $\lceil \log K\rceil$ rounds of equal length with each round $p$ using the budget  $b = \Big\lfloor \frac{T}{\lceil \log K\rceil}\Big\rfloor$. In each round $p$, the server maintains a set of active arms $A_p$ determined based on the observations received from the agents. At the beginning of the algorithm, the server assigns an index $i$ to each arm $a \in \mathcal{A}$ and sends index, arm vector pairs $(i, a_i)$ to each agent for all the arms. The server uses these indices later in the algorithm to communicate the distribution of arm pulls to the agents in each round. This saves the server from sending the actual arm vectors to the agents, thus reducing the communication overhead. For simplicity, we will denote the arms in the active set $A_p$ by $a_i$'s.\\

\noindent Throughout the execution of the algorithm, each agent $m$ maintains the local variables $V_{m}$ and $D_{m}$ to store its local reward history. Additionally, the server also maintains the local variables $V_S$ and $D_S$, used for aggregating the information communicated by the agents. At the beginning of each round $p$, the server first computes the projected vectors for the arms in the active arm set $A_p$ and determines the probability distribution $\pi_p$ over the arms by computing a 1-approximate solution to the G-optimal design problem given in Equation (\ref{g_opt}). For each arm $a_i$ in the active set $A_p$, the server computes the exact number of pulls $b_p(a_i)$ using $b_p(a_i) = b\pi_p(a_i)$ and sends the pair $(i, b_p(a_i))$ to all the agents. Each agent $m \in M$ pulls the arms as communicated by the server. After observing the corresponding rewards, the agent $m$ updates its local variables $V_m$ and $D_m$ using
\begin{equation}\label{local_update}
    V_m = \sum_{a_i \in \mathcal{A}_p} b_p(a_i)a_ia_i^{\top} \quad \text{and} \quad B_m = \sum_{a_i \in \mathcal{A}_p}\sum_{j=1}^{b_p(a_i)} a_ir_{m, j}^i.
\end{equation}
where, $r^i_{m,j}$ is the reward obtained by the agent $m$ on the $j$th pull of arm $a_i$. 

\noindent The agent $m$ sends the updated variables to the server and reset them to zero for the next round. The server then combines the local reward histories obtained from all agents and integrates them in its local variables $B_S$ and $D_S$ according to the Equation (\ref{server}). Then, the server computes the estimate $\hat{\theta}_S$ using Equation (\ref{theta_estimate}) and estimates the expected reward $x_p(a_i)$ for each arm $a_i \in \mathcal{A}_p$ using $x_p(a_i) = \hat{\theta}_S^{\top} a_i$. Finally, the server updates the active arm set for the next round to be the set of the top $K/2^p$ arms with highest estimated expected rewards $x_p(a_i)$. At this point, the server resets its local variables $B_S$ and $D_S$ for the next round:
\begin{equation}\label{server}
    V_S = \sum_{m=1}^M V_{m} \quad \text{and}\quad  D_S = \sum_{m=1}^M D_{m}.
\end{equation}

\noindent At the end of the last round, the server receives the local reward histories from all the agents and updates its local variables $B_S$ and $D_S$. Subsequently, it computes an estimate of $\theta$ utilizing these variables. The server then returns the best arm as the only arm remaining in the updated active set.

\begin{algorithm}[H]
\caption{\textsc{MaLinBAI-Star}}\label{alg:star}
\begin{algorithmic}[1]
\State \textbf{Input:} Arms set $\mathcal{A}$, budget $T$, number of agents $M$ 
\State \textbf{Initialization:} $\mathcal{A}_1 \leftarrow \mathcal{A}, b = \Big\lfloor \frac{T}{\lceil \log K\rceil}\Big\rfloor, V_S \leftarrow 0, D_S \leftarrow 0, V_{m} \leftarrow 0, D_{m} \leftarrow 0, \quad \forall m \in [M]$
\State Server assigns index and communicates the pair $(i, a_i), \forall a_i \in \mathcal{A}$ to all the agents
\For{$p = 1$ to $\lceil \log K\rceil$}
\State $d_r$ = rank($\mathcal{A}_r$) 
\State Project $\mathcal{A}_p$ to $d_p$ dimensions
\State Server computes the distribution $\pi_p$ over $\mathcal{A}_p$ by solving G-optimal design in Equation (\ref{g_opt}) with support at most $\frac{d(d+1)}{2}$
\State Server sets $b_p(a_i) = b\pi_p(a_i)$ for all $a_i \in \mathcal{A}_r$
\For{all the agents $m \in [M]$}
\State Server sends the pairs $(i, b_p(a_i)), \forall a_i \in \mathcal{A}_p$ to the agent $m$
\State Agent $m$ plays the arms following the values in the pairs $(i, b_p(a_i))$ and updates its local parameters $V_{m}, B_{m}$ using Equation (\ref{local_update}).
\State Agent $m$ sends $V_{m}, B_{m}$ to the server and resets them for next round.
\EndFor
\State Server updates its local parameters $V_S$ and $B_S$ using Equation (\ref{server}) and computes the estimate $\hat{\theta}_S$ using Equation (\ref{theta_estimate}).
\For{$a_i \in \mathcal{A}_p$}
\State Server computes the estimated expected reward $x_p(a_i) = \hat{\theta}_S^{\top}a_i$.
\EndFor
\State Server computes the set $S_p$ of top $K/2^p$ arms with highest estimated expected reward and sets $\mathcal{A}_{p+1} = S_p$
\State Server resets $B_S$ and $D_S$.
\EndFor
\State Server outputs the only arm in $\mathcal{A}_{\lceil \log K\rceil}$ as the best arm.
\end{algorithmic}
\end{algorithm}

\noindent \textbf{Communication cost:} Our algorithm requires $\lceil\log K\rceil$ rounds of communication between agents and server. Each round involves exchange of a data message between $M$  agents and the central server. Therefore, total communication cost incurred by our algorithm is $O(2M\log K)$.

\begin{lemma}\label{lemma1}
    Let $a_1$ be the best arm, and suppose that $a_1$ is not eliminated prior to round $p$, the probability that any suboptimal arm $a_i$ has estimated expected reward higher than that of $a_1$ during round $p$ of \textsc{MaLinBAI-Star} is given by
    \[\Pr[x_p(a_i) > x_p(a_1)] \leq \exp\left\{\frac{-TM\Delta_i^2}{16d\lceil \log K\rceil} \right\}.\]
\end{lemma}
\noindent This lemma plays a critical role in establishing the theoretical guarantees of our algorithms. In particular, we use it in Lemma \ref{lemma2} to bound the probability of eliminating the best arm during the execution of \textsc{MaLinBAI-Star}. The proof of this lemma is mainly driven by a concentration bound for the sequence of arm pulls \cite{lattimore2020bandit}, and the properties of G-optimal design. 

\begin{lemma}\label{lemma2}
    Let $a_1$ be the best arm, and suppose that $a_1$ is not eliminated prior to round $p$, the probability that the best arm is eliminated in round $p$ of \textsc{MaLinBAI-Star} is given by
    \[\Pr[a_1 \notin \mathcal{A}_{p+1}|a_1 \in \mathcal{A}_{p}] \leq 2 \exp\left\{\frac{-TM\Delta_{\min,p}^2}{16d\lceil \log K\rceil} \right\},\]
    where $\Delta_{\min,p}$ is the minimum sub-optimality gap of the active arm set $\mathcal{A}_{p}$.
\end{lemma}
\noindent The proof of this lemma relies on applying Markov's inequality on the expected number of arms having higher estimated expected reward than that of the best arm. Detailed proofs of Lemma \ref{lemma1} and Lemma \ref{lemma2} are given in Appendix B.

\begin{theorem}\label{theorem1}
    Given time budget $T$, the probability of error in estimating the best arm by \textsc{MaLinBAI-Star} is given by
    \[\Pr[Error] \leq 4\log K \exp\left\{\frac{-TM\Delta_{\min}^2}{32d\log K} \right\},\]
    where $\Delta_{\min}$ is the minimum sub-optimality gap of the given problem instance.
\end{theorem}
\begin{proof} Notice that
    \begin{align*}
        \Pr[Error] & = \Pr[a_1 \notin \mathcal{A}_{\lceil \log K\rceil +1}]\\
        & \leq \sum_{p=1}^{\lceil \log K\rceil} \Pr[a_1 \notin \mathcal{A}_{p+1}|a_1 \in \mathcal{A}_{p}]\\
        & \leq \sum_{p=1}^{\lceil \log K\rceil} 2 \exp\left\{\frac{-TM\Delta_{\min,p}^2}{16d\lceil \log K\rceil} \right\} \hspace{4.95cm} (\text{using Lemma \ref{lemma2}})\\
        & \leq \sum_{p=1}^{\lceil \log K\rceil} 2 \exp\left\{\frac{-TM\Delta_{\min}^2}{16d\lceil \log K\rceil} \right\} \hspace{4.2cm} (\text{using $\Delta_{\min,p} \geq \Delta_{\min}$})\\
        & = 2 \lceil \log K\rceil \exp\left\{\frac{-TM\Delta_{\min}^2}{16d\lceil \log K\rceil} \right\} \\&\leq 4\log K \exp\left\{\frac{-TM\Delta_{\min}^2}{32d\log K} \right\},
    \end{align*} as desired.
\end{proof}

\section{Algorithm for Generic Networks}\label{section:gen}
This section details our algorithm for generic networks. We consider multiple agents connected through a network represented by a graph $G = (V, E)$, where each vertex of $G$ represent an agent, and an edge between two vertices $u$ and $v$ represent the connection between the corresponding agents.\\

\noindent Given such a graph $G$, our algorithm \textsc{MaLinBAI-Gen} first partitions the graph into dominating sets. For this purpose, we first state following definitions:\\

\noindent \textit{Dominating Set:} Given a graph $G = (V, E)$, a \emph{dominating set} $D$ of $G$ is a subset of $V$ such that for any vertex $u \in V\setminus D$, there is a vertex $v \in D$ such that $v$ is connected to $u$ by an edge.\\

\begin{figure}[!ht]
\begin{minipage}{0.49\textwidth}
    \centering
    \includegraphics[width=0.75\textwidth]{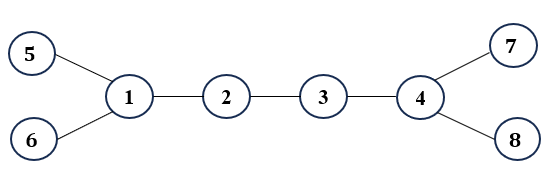}
    \subcaption{Example graph with 8 nodes}
    \end{minipage}
\begin{minipage}{0.49\textwidth}
    \centering
    \includegraphics[width=0.7\textwidth]{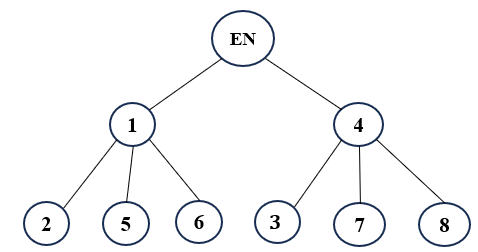}
    \subcaption{Corresponding 2-tier star network structure from the dominating set partition $\{P_1, P_2\} = \{\{1,2,5,6\},\{3,4,7,8\}\}$ arising from the dominating set $\{1,4\}$.}
    \end{minipage}
    \caption{Illustration of two-tier star network}
    \label{fig:network_structure}
\end{figure}

\noindent \textit{Dominating Set Partition:} Given a graph $G$ and a dominating set $D$, the corresponding \emph{dominating set partition} $P=\{P_1,\ldots,P_{|D|}\}$ is obtained by partitioning the vertices of $G$ into $|D|$ subsets (referred to as \emph{blocks} or \emph{parts}) such that each $P_i$ (with $1\leq i\leq |D|$)  contains exactly one vertex from $v\in D$, known as the dominant vertex, and a subset of vertices directly connected to $v$ by an edge (see Figure~\ref{fig:network_structure}).\\ 

\noindent The pseudocode for our algorithm \textsc{MaLinBAI-Gen} is presented in Algorithm~\ref{alg:gen}. This algorithm is a natural extension of \textsc{MaLinBAI-Star} to arbitrary networks. Given a network $G = (V, E)$ and a dominating set partition $P$ of $G$, each block $P_i$ in the partition $P$ represents a star network, where the corresponding dominant vertex $p_i^h$ plays the role of central server or hub. So each block $P_i$ executes \textsc{MaLinBAI-Star} on the given bandit instance with the variation that the central hub $p_i^h$ also participates in the learning process by playing the actions and maintaining its local reward history.\\

\noindent In this structure, there is a top-level server which is responsible for estimating the best arm at the end of the budget. The ensembler directly communicates with the hub nodes of partition $P$. This configuration forms a hierarchical two-level star network as shown in Figure \ref{fig:network_structure}. At the first level, the server is linked to $|P|$ hub nodes, each of which is further connected to a subset of $V$. \\

\noindent Until the end of the budget, each block $P_i$ executes \textsc{MaLinBAI-Star} independently where the hub nodes $p_i^h$ also play the actions and maintain their local reward history. At the end of the budget, the top-level server receives the estimated best arm from the hub node of each block in the partition $P$. Then the server determines the global best arm my using the majority vote mechanism. Specifically, the server selects the arm that appears most frequently among the $|P|$ best arms chosen by each hub node as the final best arm. In the event of a tie, i.e., multiple arms receiving the highest number of votes, the server determines the final best arm by selecting the arm with the lowest uncertainity in its estimated expected reward among the contending arms. For any arm $a$, this uncertainty is quantified using the expression $a^{\top}V_p^{-1}a$, which reflects the confidence of each hub in its estimated expected reward for the selected arm.

\begin{algorithm}[H]
\caption{\textsc{MaLinBAI-Gen}}\label{alg:gen}
\begin{algorithmic}[1]
\State \textbf{Input:} Arms set $\mathcal{A}$, budget $T$, a dominating set partition $P = \{P_1, P_2, \ldots, P_{|D|}\}$ on network graph $G = (V, E)$
\State \textbf{Initialization:} $\mathcal{A}_1 \leftarrow \mathcal{A}, b = \Big\lfloor \frac{T}{\lceil \log K\rceil}\Big\rfloor, V_S \leftarrow 0, D_S \leftarrow 0, V_{m} \leftarrow 0, D_{m} \leftarrow 0, \quad \forall m \in V$
\For{$P_i \in P$}
\For{$p = 1$ to $\lceil \log K\rceil$}
\State Execute lines (5)-(8) of \textsc{MaLinBAI-Star} 
\For{$m \in P_i$}
\State Execute lines (10)-(12) of \textsc{MaLinBAI-Star}
\EndFor
\State Hub updates its local variables
\State $V_{p_i^h} \leftarrow \sum_{m\in P_i} V_{m}$
\State $D_{p_i^h} \leftarrow \sum_{m\in P_i} D_{m}$
\State $\hat{\theta}_{p_i^h} = V_{p_i^h}^{-1}D_{p_i^h}$
\State Execute lines (15)-(19) of \textsc{MaLinBAI-Star}
\EndFor
\State Return the only arm in $\mathcal{A}_{\lceil \log K\rceil}$ and it's corresponding variance.
\EndFor
\State The server receives the estimated best arm and it's variance from all hub nodes
\State Find most frequently chosen best arm $a_{out}$ among $|P|$ hub nodes
\If{$a_{out}$ is unique}
\State Return $a_{out}$
\Else{}
\State Return the arm with lowest variance among possible values of $a_{out}$
\EndIf
\end{algorithmic}
\end{algorithm}

\noindent \textbf{Communication cost:} The execution of \textsc{MaLinBAI-Star} for each $P_i \in P$ requires $O(2M_i\log K)$ communication cost, where $M_i$ is the number of agents present in $P_i\in P$ other than the dominant vertex $p_i^h$. \textsc{MaLinBAI-Gen} executes \textsc{MaLinBAI-Star} for each $P_i \in P$ which takes total $O(\sum_{i=1}^{|P|}O(2M_i\log K)) = O(2(M - |P|)\log K)$ cost. At the end, all hub nodes send their estimated best arm to ensemble which adds $|P|$ to the cost. Therefore, the total communication cost required by \textsc{MaLinBAI-Gen} is $O(2(M - |P|)\log K + |P|)$. 


\begin{theorem}\label{theorem2}
    Given time budget $T$, the probability of error in estimating the best arm by \textsc{MaLinBAI-Gen} is given by
    \[\Pr[Error] \leq 8\log K \exp\left\{\frac{-T\Delta_{\min}^2}{32d\log K} \right\},\]
    where $\Delta_{\min}$ is the minimum sub-optimality gap of the given bandit instance.
\end{theorem}

\noindent It is important to note that this upper bound on error probability is independent of the size of the dominating set partition of the underlying graph. This highlights the fact that our algorithm \textsc{MaLinBAI-Gen} does not require finding a minimum dominating set partition of the given graph—a problem known to be NP-Hard.\\

\noindent The proof of Theorem \ref{theorem2} primarily builds upon an application of Markov's inequality, combined with Theorem \ref{theorem1}. A complete proof is provided in Appendix C.\\

\noindent\textit{\textbf{Remark:} In Appendix \ref{appendix:D}, we show that the error probability of our algorithms matches the lower bound in \cite{yang2022minimax} up to a factor of $(\log d/\log K)$. Additionally, the communication cost of our algorithms is shown to match the lower bound presented in \cite{he2022simple}, up to a $(\log T\log K)$ factor difference. As a result, our algorithms achieve near-optimal error rates while maintaining efficient communication costs.}

\section{Experiments}\label{sec:experiment}

In this section, we compare the performance of our algorithm \textsc{MaLinBAI-Star} with several baselines including including (1) minimax optimal single-agent fixed-budget pure-exploration algorithm OD-LinBAI \cite{yang2022minimax}, (2) multi-agent regret minimization algorithms FedLinUCB\cite{he2022simple}, Async-LinUCB \cite{li2022asynchronous}, and (3) multi-agent fixed-confidence best-arm identification algorithm FALinPE \cite{wang2023pure}.\\

\noindent We introduce a variant of OD-LinBAI, called MA-OD-LinBAI, to adapt its functionality for multi-agent environments. In this variant, with $M$ agents, MA-OD-LinBAI independently runs OD-LinBAI $M$ times, and the final best arm is selected as the one that appears most frequently among the $M$ best arms chosen by each agent. For regret-minimization algorithms, the best arm is identified as the arm pulled most frequently by the agents by the end of the allocated budget, with ties broken randomly. In the case of fixed-confidence algorithm FALinPE, we apply the stopping rule at the end of the budget to finalize the best arm.\\

\begin{figure}[!ht]
  \begin{minipage}{0.49\textwidth}
    \centering
    \includegraphics[width=\textwidth]{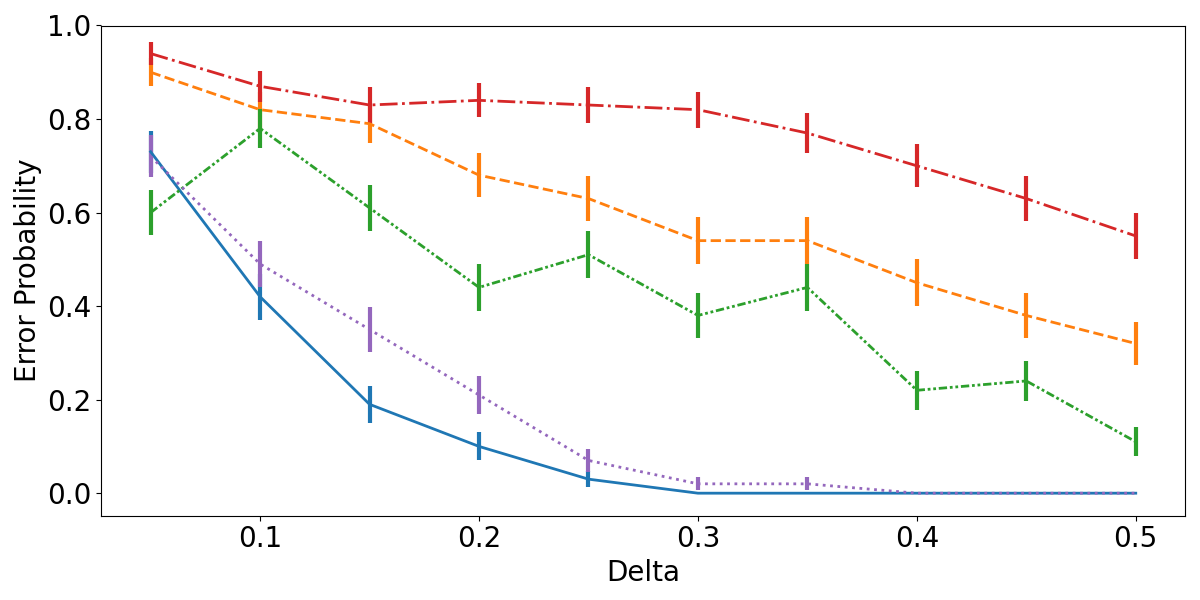}
    \subcaption{Standard instance}
    \label{fig:delta_comp}
  \end{minipage}
  \begin{minipage}{0.49\textwidth}
    \centering
    \includegraphics[width=\textwidth]{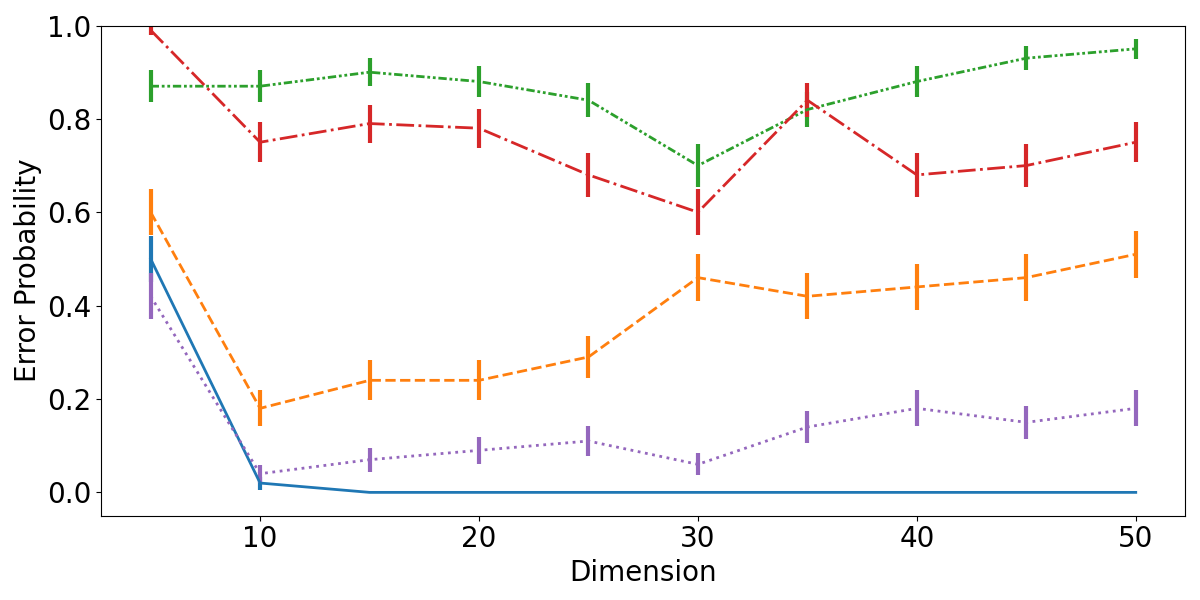}
    \subcaption{Random instance}
    \label{fig:sphere_comp}
  \end{minipage}
    \centering
    \includegraphics[width=0.8\textwidth]{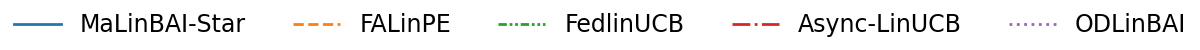}
    \label{fig:legend}
  \caption{Experimental results for synthetic data.}
  \label{comparison}
\end{figure}

\noindent We evaluate the algorithms on both the real-world MovieLens dataset \cite{harper2015movielens} and two types of synthetic datasets: (1) Standard instance and, (2) Random instance. For each of the instance, we execute the algorithms for time budget $150$ and number of agents $M = 15$. Noise is set as $\eta \sim \mathcal{N}(0,1)$ generated independently for each trial. We tune the parameter $\alpha$ for FedLinUCB and $\gamma_U=\gamma_D$ for Async-LinUCB, setting $\alpha=1$ and $\gamma_U=\gamma_D=5$ to match the optimal performance reported in the original papers. For FALinPE, we adopt the parameter values shown to achieve the lower bounds, specifically, $\gamma_1 = 1/M^2, \gamma_2 = 1/(2MK)$. The probability of error for each experiment is calculated as the fraction of errors made by the algorithms over 100 Monte Carlo simulations. The standard error of error probabilities is represented by the tiny error bars in Figure \ref{comparison}. Due to space constraints, further discussion on numerical analysis and the results observed for real-world dataset are provided in Appendix E.\\

\noindent \textbf{Standard Instance:} In this setup, the arm set consists of $d$-dimensional canonical vectors $\{e_1, e_2, \ldots, e_d\}$ with $d=10$. The unknown vector $\theta$ is set to $\Delta e_1$, where $\Delta > 0$ thus, establishing first arm $e_1$ as the best arm. We execute the algorithms for $\Delta$ values ranging from $0.05$ to $0.5$, $\Delta = \{0.05, 0.1, \ldots, 0.5\}$.\\

\noindent \textbf{Random Instance:} In this dataset, the arm set $\mathcal{A}$ consists of 100 random vectors drawn from the unit sphere in $\mathbb{R}^{d-1}$ centered at the origin. To construct the unknown vector $\theta$, we choose two closest vectors from $\mathcal{A}$. Without loss of generality, let's assume that $u$ and $v$ are the two closest vectors in $\mathcal{A}$, we then define $\theta$ to be $u + 0.01(u - v)$, thereby designating $u$ as the optimal arm. Within this framework, we evaluate the algorithms for the dimension ranging from 5 to 50.

\noindent In Figure \ref{comparison}, we observe that \textsc{MaLinBAI-Star} consistently outperforms MA-OD-LinBAI, Async-LinUCB, FedLinUCB, and FALinPE across all datasets in terms of best arm identification. In Appendix E, we present the numerical results on the communication cost required by each of the algorithms. These results show that, while the communication cost incurred by \textsc{MaLinBAI-Star} is only marginally higher than that of Async-LinUCB and FedLinUCB, it is significantly lower than the cost required by FALinPE. However, none of these algorithms perform better than our algorithm \textsc{MaLinBAI-Star} in accurately estimating the best arm. Thus, \textsc{MaLinBAI-Star} strikes an effective balance by achieving both near-optimal error probability and efficient communication costs for federated linear bandits.

\section{Conclusion and Future Work}\label{sec:conclusion}
In this paper, we study the problem of fixed-budget, best-arm identification for stochastic linear bandits in a multi-agent setting. We devise two algorithms \textsc{MaLinBAI-Star} and \textsc{MaLinBAI-Gen}, which consider a network of agents connected through a star network or a generic network. The agents collaborate to learn the best arm with the help of a central server. \textsc{MaLinBAI-Gen} takes into account the structure of the network by using dominating set partitions. Both algorithms utilize the technique of G-optimal design along with the successive elimination based strategy and achieve exponentially decaying error probability in the allocated time budget.\\ 

\noindent As for future work, one could study a decentralized setting where agents need not be dependent on a central server for information sharing. Furthermore, designing a fault-tolerant, fully distributed algorithms is also a reasonable direction of study since in practice it is possible that agents may intermittently go offline.



\newpage

\appendix

\section{Notations and Technical Lemmas}
\textbf{Notation:} If $n$ is a positive integer, $[n]$ denotes the set $\{1,2,\ldots,n\}$. Furthermore, if $x\in\mathbb{R}^d$, then $\|x\|_2$ denotes the 2-norm of $x$ and $\|x\|_A$ denotes $\sqrt{x^{\top} Ax}$. Finally, if $C,D$ are sets, $|C|$ denotes the cardinality of $C$ and $C\setminus D$ denotes their set difference. 

Next, we present some key lemmas crucial for deriving the theoretical guarantees of our algorithms,

\begin{lemma}[Kiefer and Wolfowitz\cite{kiefer1960equivalence}]\label{lemma:g-opt}
    Given a set of d-dimensional vectors $\mathcal{A} \subset \mathbb{R}^d$, suppose that $\mathcal{A}$ spans $\mathbb{R}^d$. Then, there exists a probability distribution $\pi^*: \mathcal{A} \rightarrow [0,1]$ with $\sum_{a \in \mathcal{A}} \pi(a) = 1$ such that following statements are equivalent:
    \begin{enumerate}
        \item $\pi^*$ is a minimizer of $g$,
        \item $\pi^*$ is a maximizer of $f(\pi) = \log \det V(\pi)$, and
        \item $g(\pi^*) = d$.
    \end{enumerate}
    Where
    \[ g(\pi) = \max_{a \in \mathcal{A}}\|a\|_{V(\pi)^{-1}}^2, \quad \text{with} \quad V(\pi) = \sum_{a \in \mathcal{A}} \pi(a)aa^{\top}\]
    Furthermore, there exists a minimiser $\pi^*$ of $g$ such that $|Supp(\pi^*)| \leq d(d + 1)/2$.
\end{lemma}

\begin{lemma}[Lattimore and Szepesv\'ari\cite{lattimore2020bandit}]\label{lemma:concentration-bound}
    Suppose a bandit algorithm has chosen actions $A_1, A_2, ..., A_t \in \mathbb{R}^d$ and received the rewards $X_1, X_2, ..., X_t$ with $X_s = \left<\theta_*, A_s \right> + \eta_s$ where $\eta_s$ is independent subgaussian noise. If $A_1, A_2, ..., A_t$ are deterministically chosen without the knowledge of $X_1, X_2, ..., X_t$, then for any vector $a \in \mathbb{R}^d$ and $\delta > 0$,
    \[\Pr[\left<\hat{\theta}_t - \theta_*, a \right> \geq \sqrt{2\|a\|_{V_t^{-1}}^2 \log\left(\frac{1}{\delta} \right)}] \leq \delta,\]
    where, $\hat{\theta}_t$ is the OLS estimate of $\theta_*$ and $V_t = \sum_{s=1}^t A_sA_s^T$.
\end{lemma}

\begin{lemma}[Markov's Inequality]\label{lemma:Markov}
    If $X$ is a nonnegative random variable with with finite expectation $\mathbb{E}[X] < \infty$, then for any $a > 0$, the probability that $X$ is at least $a$ is at most the expectation of $X$ divided by $a$
    \[\Pr[X \geq a] \leq \frac{\mathbb{E}[X]}{a}.\]
\end{lemma}

\begin{lemma}\label{lemma:norm}
    At the end of any round $p$ during the execution of \textsc{MaLinBAI-Star}, after the server aggregates the matrices $V_m$ received from all agents into its local matrix $V_S$, the following bound holds for any arm $a_i \in \mathcal{A}_p$:
    \[\|a_i\|_{V_S^{-1}}^2 = \frac{\lceil \log K \rceil}{TM}\|a_i\|_{V(\pi_p)^{-1}}^2,\]
where, $V(\pi_p)$ denotes the design matrix corresponding to the arm pulls allocated in round $p$.
\end{lemma}
\begin{proof} Notice that
    \begin{align*}
        \|a_i\|_{V_S^{-1}}^2 = a_i^{\top}V_S^{-1}a_i & = a_i^{\top}\left(\sum_{m=1}^M \sum_{a_j \in \mathcal{A}_p} b_p(a_j)a_ja_j^{\top} \right)^{-1}a_i \\
        & = a_i^{\top}\left(\sum_{m=1}^M \sum_{a_j \in \mathcal{A}_p} b\pi_p(a_j)a_ja_j^{\top} \right)^{-1}a_i\\
        & = a_i^{\top}\left(\sum_{m=1}^M b \sum_{a_j \in \mathcal{A}_p} \pi_p(a_j)a_ja_j^{\top} \right)^{-1}a_i\\
        & = a_i^{\top}\left(\sum_{m=1}^M b V(\pi_p)\right)^{-1}a_i\\
        & = \frac{1}{bM}a_i^{\top}V(\pi_p)^{-1}a_i\\
        & = \frac{\lceil \log K \rceil}{TM}\|a_i\|_{V(\pi_p)^{-1}}^2,
    \end{align*} as claimed.
\end{proof}

\section{Missing proofs of Section \ref{section:star}}\label{appendix:B}
\subsection{Proof of Lemma \ref{lemma1}}
\begin{lemmarestate}
    Let $a_1$ be the best arm, and suppose that $a_1$ is not eliminated prior to round $p$, the probability that any suboptimal arm $a_i$ has estimated expected reward higher than that of $a_1$ during round $p$ of \textsc{MaLinBAI-Star} is given by
    \[\Pr[x_p(a_i) > x_p(a_1)] \leq \exp\left\{\frac{-TM\Delta_i^2}{16d\lceil \log K\rceil} \right\}.\]
\end{lemmarestate}
\begin{proof}
Let's assume that $\theta_*$ is the true value of the parameter $\theta$ and  w.l.o.g. assume that $a_1$ is the best arm. Let $V_{p,S}$ be the local matrix and $\hat{\theta}_{p,S}$ be the estimate calculated by the server at the end of round $p$.
    \begin{align}
        \Pr[x_p(a_i) > x_p(a_1)] & = \Pr[\left<\hat{\theta}_{p,S},a_i\right> > \left<\hat{\theta}_{p,S},a_1\right>] \\
        & = \Pr[\left<\hat{\theta}_{p,S},a_i\right> +\left<\theta_*,a_1\right> - \left<\theta_*,a_i\right> > \left<\hat{\theta}_{p,S},a_1\right> + \left<\theta_*,a_1\right> - \left<\theta_*,a_i\right>] \\
        & = \Pr[\left<\hat{\theta}_{p,S},a_i\right> - \left<\hat{\theta}_{p,S},a_1\right> + \left<\theta_*,a_1\right> - \left<\theta_*,a_i\right> > \left<\theta_*,a_1\right> - \left<\theta_*,a_i\right>] \\
        & = \Pr[\left<\hat{\theta}_{p,S} - \theta_*, a_i - a_1 \right> > \Delta_i] \\
        & \leq \exp{\frac{-\Delta_i^2}{2\|a_i-a_1\|_{V_{p,S}^{-1}}^2}}\label{B.5}\\
        & \leq \exp{\frac{-\Delta_i^2}{8\max_{a_i \in \mathcal{A}_p}\|a_i\|_{V_{p,S}^{-1}}^2}}\label{B.6}\\
        & \leq \exp{\frac{-TM\Delta_i^2}{8\lceil \log K \rceil\max_{a_i \in \mathcal{A}_p}\|a_i\|_{V(\pi_p)^{-1}}^2}}\label{B.7}\\
        & \leq \exp{\frac{-TM\Delta_i^2}{16\lceil \log K \rceil d}},\label{B.8} \end{align}

        \noindent where we used Lemma \ref{lemma:concentration-bound} in (\ref{B.5}), and (\ref{B.6}) follows from
    \[\|a_i-a_1\|_{V_{p,S}^{-1}} \leq \|a_i\|_{V_{p,S}^{-1}} + \|a_1\|_{V_{p,S}^{-1}} \leq 2\max_{a_i \in \mathcal{A}_p}\|a_i\|_{V_{p,S}^{-1}}.\]
    (\ref{B.7}) uses Lemma \ref{lemma:norm} and (\ref{B.8}) follows from Lemma \ref{lemma:g-opt}.
\end{proof}

\subsection{Proof of Lemma \ref{lemma2}}
\begin{lemmarestate}
    Let $a_1$ be the best arm, and suppose that $a_1$ is not eliminated prior to round $p$, the probability that the best arm is eliminated in round $p$ of \textsc{MaLinBAI-Star} is given by
    \[\Pr[a_1 \notin \mathcal{A}_{p+1}|a_1 \in \mathcal{A}_{p}] \leq 2 \exp\left\{\frac{-TM\Delta_{\min,p}^2}{16d\lceil \log K\rceil} \right\},\]
    where $\Delta_{\min,p}$ is the minimum sub-optimality gap of the active arm set $\mathcal{A}_{p}$.
\end{lemmarestate}
\begin{proof}
    Let $I_p(a_i)$ be the indicator random variable corresponding to the event that the estimated expected reward of arm $a_i$ is higher than that of arm $a_1$ at the end of round $p$. Formally,
    \[I_p(a_i) = \begin{cases}
                1 & if \hspace{0.2cm} x_p(a_i) > x_p(a_1)\\
                0 & otherwise
\end{cases}\]
Let $N_p$ be the random variable denoting the number of suboptimal arms having estimated expected reward higher than that of arm $a_1$ at the end of round $p$, i.e., $N_p = \sum_{a_i \in \mathcal{A}_p} I_p(a_i)$.
\begin{align}
    \mathbb{E}[N_p] & = \mathbb{E}\left[\sum_{a_i \in \mathcal{A}_p} I_p(a_i)\right] \\
    & = \sum_{a_i \in \mathcal{A}_p} \Pr[x_p(a_i) > x_p(a_1)]\label{B.10} \\
    & \leq \sum_{a_i \in \mathcal{A}_p} \exp{\frac{-TM\Delta_i^2}{16\lceil \log K \rceil d}}\label{B.11}\\
    & \leq \sum_{a_i \in \mathcal{A}_p} \exp{\frac{-TM\Delta_{\min,p}^2}{16\lceil \log K \rceil d}}\label{B.12} \\
    & \leq |\mathcal{A}_p| \exp{\frac{-TM\Delta_{\min,p}^2}{16\lceil \log K \rceil d}},
\end{align}
where (\ref{B.10}) follows from the properties of indicator random variable, (\ref{B.11}) uses Lemma \ref{lemma1}, and in (\ref{B.12}) we used that $\Delta_i \geq \min_{a_i \in \mathcal{A}_p} \Delta_{a_i} = \Delta_{\min,p}$.\\

\noindent Note that, the best arm $a_1$ gets eliminated in round $p$ only if more than half of the arms in the active arm set $\mathcal{A}_p$ have estimated expected reward higher than that of arm $a_1$ at the end of round $p$. \\

\noindent Now using Markov's inequality (Lemma \ref{lemma:Markov}),
\begin{align*}
    \Pr\left[N_p > \frac{|\mathcal{A}_p|}{2}\right] & < \frac{2\mathbb{E}[N_p]}{|\mathcal{A}_p|}\\
    & \leq \frac{2}{|\mathcal{A}_p|}|\mathcal{A}_p| \exp{\frac{-TM\Delta_{\min,p}^2}{16\lceil \log K \rceil d}}\\
    & = 2\exp{\frac{-TM\Delta_{\min,p}^2}{16\lceil \log K \rceil d}}, 
\end{align*} as claimed.
\end{proof}

\section{Missing proofs of Section \ref{section:gen}}\label{appendix:C}
\subsection{Proof of Theorem \ref{theorem2}}
\begin{theoremrestate}
    Given time budget $T$, the probability of error in estimating the best arm by \textsc{MaLinBAI-Gen} is given by
    \[\Pr[Error] \leq 8\log K \exp\left\{\frac{-T\Delta_{\min}^2}{32d\log K} \right\},\]
    where $\Delta_{\min}$ is the minimum sub-optimality gap of the given bandit instance.
\end{theoremrestate}
\begin{proof}
    Given the network graph $G = (V, E)$, let $P$ be a dominating set partition of $G$, and $P_1, P_2, ..., P_{|P|}$ be the blocks of $P$. Also, w.l.o.g. assume that $a_1$ is the best arm and the best arm estimated on $P_i$ is denoted by $a^*(P_i)$.\\
    
    \noindent Let $I(P_i)$ be the indicator random variable corresponding to the event that the best arm estimated on $P_i$ is not $a_1$. Formally,
    \[I_p(a_i) = \begin{cases}
                1 & if \hspace{0.2cm} a^*(P_i) \neq a_1\\
                0 & otherwise
\end{cases}\]
Let $D_P$ be the random variable denoting the number of blocks in the partition $P$ having the estimated best arm other than $a_1$, i.e., $D_P = \sum_{P_i \in P} I(P_i)$. Let $\Pr[Error_{P_i}]$ denote the probability of making error in identifying the best arm by the agents in $P_i$ and $M_{P_i}$ be the number of agents in $P_i$.
\begin{align}
    \mathbb{E}[D_P] & = \mathbb{E}\left[\sum_{P_i \in P} I(P_i)\right] \\
    & = \sum_{P_i \in P} \Pr[Error_{P_i}]\label{C.2} \\
    & \leq \sum_{P_i \in P} 4\log K \exp\left\{\frac{-TM_{P_i}\Delta_{\min}^2}{32d\log K} \right\}\label{C.3}\\
    & \leq \sum_{P_i \in P} 4\log K \exp\left\{\frac{-T\Delta_{\min}^2}{32d\log K} \right\}\label{C.4} \\
    & \leq |P| 4\log K \exp\left\{\frac{-T\Delta_{\min}^2}{32d\log K} \right\},
\end{align}
where (\ref{C.3}) follows from Theorem \ref{theorem1} which gives the probability of making error in identifying the best arm by \textsc{MaLinBAI-Star} on star graph with $M$ agents. In (\ref{C.4}), we used the fact that each block $P_i \in P$ has at least one agent, that is, $M_{P_i} \geq 1, \forall i \in [|P|]$.\\

\noindent Note that the server fails to correctly identify the best arm as $a_1$ either when a strict majority of blocks in $P$ select a suboptimal arm as the best arm, or when a tie occurs and the tie-breaking mechanism selects an arm other than $a_1$. In both the cases, it must be that strictly more than half of the blocks $P_i$ ($1\leq i\leq |D|$) fail to identify $a_1$ as the best arm. Therefore, bounding the probability that $D_P > |P|/2$ suffices to capture the total probability of error made by the server, covering both majority vote and tie-breaking scenarios.\\ 

\noindent Now using Markov's inequality (Lemma \ref{lemma:Markov}),
\begin{align*}
    \Pr\left[D_P > \frac{|P|}{2}\right] & < \frac{2\mathbb{E}[D_P]}{|P|}\\
    & \leq \frac{2}{|P|} |P| 4\log K \exp\left\{\frac{-T\Delta_{\min}^2}{32d\log K} \right\}\\
    & = 8\log K \exp\left\{\frac{-T\Delta_{\min}^2}{32d\log K} \right\},
\end{align*} as desired. 
\end{proof}

\section{Proof of Lower Bound}\label{appendix:D}
\subsection{Lower Bound for Error Probability}
According to Theorem 3 of \cite{yang2022minimax}, the lower bound on the probability of error in estimating the best arm is given by
\begin{equation}\label{lower_bound}
    \exp\left(-O\left(\frac{T}{H_{1,lin(v)}\log_2 d}\right)\right),
\end{equation}
where $H_{1,lin(v)} = \sum_{1 \leq i \leq d} \Delta_i^{-2}$ denotes the hardness quantity of bandit instance $v$.
\[\sum_{1 \leq i \leq d} \Delta_i^{-2} \leq \sum_{1 \leq i \leq d} \Delta_{\min}^{-2}
    = \frac{d}{\Delta_{\min}^{2}}\]
Therefore, quantity (\ref{lower_bound}) can be re-written as,
\[\exp\left(-O\left(\frac{T}{H_{1,lin(v)}\log_2 d}\right)\right) = \exp\left(-O\left(\frac{\Delta_{\min}^{2}T}{d\log d}\right)\right).\]

\noindent The error probability of our algorithm \textsc{MaLinBAI-Star} is given by
\[\Pr[Error] \leq 4\log K \exp\left\{\frac{-TM\Delta_{\min}^2}{32d\log K} \right\},\]

\noindent For the single agent case i.e., $M=1$, error probability of \textsc{MaLinBAI-Star} can be written as
\begin{align*}
    \Pr[Error] \leq 4\log K \exp\left\{\frac{-T\Delta_{\min}^2}{32d\log K} \right\} = \exp\left\{-O\left(\frac{\Delta_{\min}^{2}T}{d\log K}\right)\right\}.
\end{align*}
Therefore, the upper bound for the error probability of our algorithm \textsc{MaLinBAI-Gen} matches the lower bound in \cite{yang2022minimax} up to a factor of $\log d/\log K$.

\subsection{Lower Bound for Communication cost}
According to Theorem 5.3 of \cite{he2022simple}, the lower bound on the communication cost in a $M$ node network is given by $O(M/\log(T/M))$. This lower bound can be further simplified to $O(M/\log T)$.\\

\noindent The communication cost of our algorithm \textsc{MaLinBAI-Star} is given by $O(2M\lceil \log K\rceil)$. This can be expressed as $O(M\log K)$ which is $(\log T\log K)$ away from the lower bound. Therefore, we conclude that our algorithm matches the lower bound for communication cost presented in \cite{he2022simple} up to a factor of $(\log T\log K)$.\\

\noindent It is important to highlight that the lower bound discussed in \cite{he2022simple} pertains to regret minimization, whereas our algorithms focus on the task of best-arm identification. However, the work in \cite{he2022simple} also addresses a multi-agent framework for linear bandits, which is closely aligned with the setup we consider in our work.

\section{Additional Experiments}\label{appendix:E}
\subsection{Communication Cost}\label{exp:comm_cost}
\begin{figure}[!ht]
  \begin{minipage}{0.48\textwidth}
    \centering
    \includegraphics[width=\textwidth]{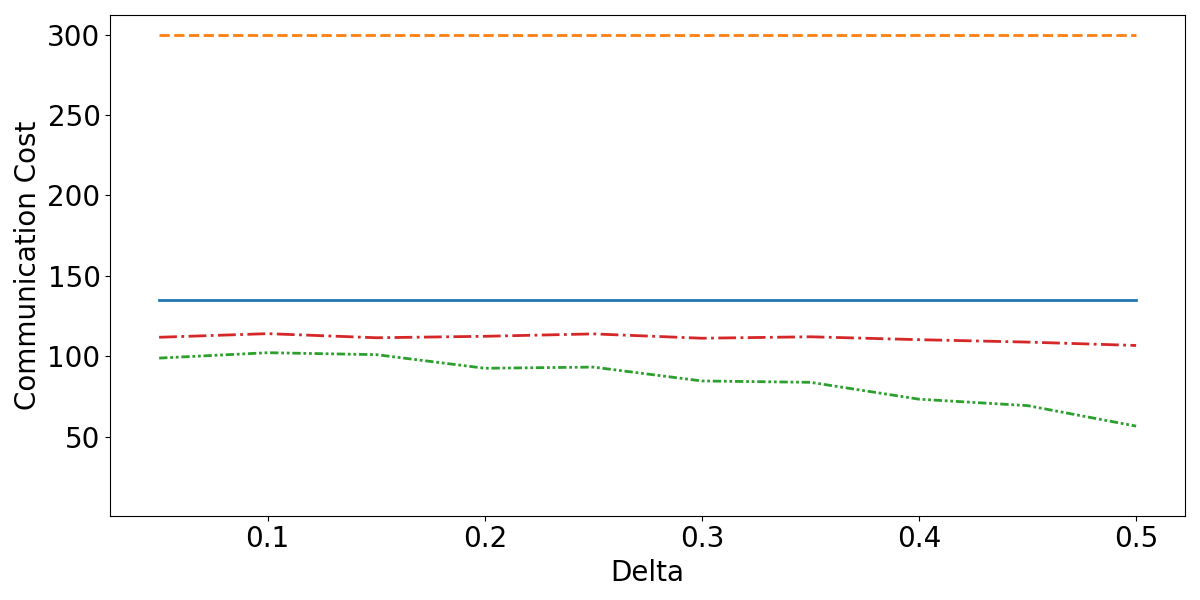}
    \subcaption{Standard instance}
    \label{fig:delta_cost_comp}
  \end{minipage}
  \begin{minipage}{0.48\textwidth}
    \centering
    \includegraphics[width=\textwidth]{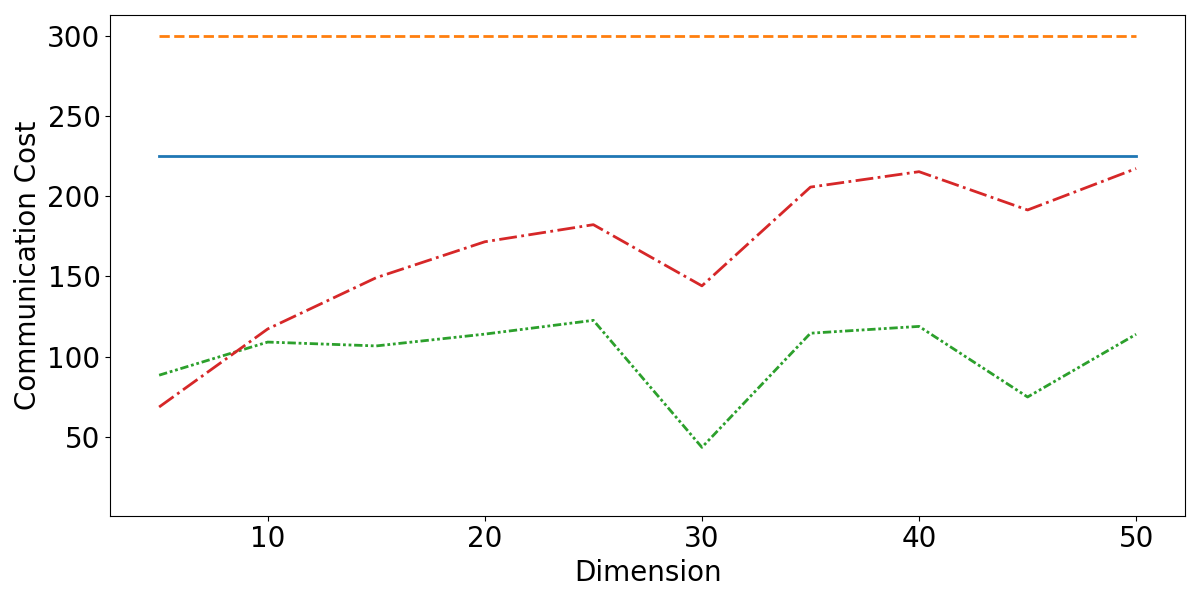}
    \subcaption{Random instance}
    \label{fig:sphere_cost_comp}
  \end{minipage}
  \centering
    \includegraphics[width=0.8\textwidth]{legend.png}
  \caption{Communication cost experimental results for synthetic data.}
  \label{app_comm_cost}
\end{figure}

In this section, we present the numerical results for the communication cost incurred by the algorithms under the experiments performed in Section 5. Recall that, we execute the algorithms on two types of synthetic datasets: (1) Standard instance and (2) Random instance. For each of the instance, we execute the algorithms for time budget $150$ and number of agents $M = 15$. Noise is set as $\eta \sim \mathcal{N}(0,1)$ generated independently for each trial. We tune the parameter $\alpha$ for FedLinUCB and $\gamma_U=\gamma_D$ for Async-LinUCB, setting $\alpha=1$ and $\gamma_U=\gamma_D=5$ to match the optimal performance reported in the original papers. For FALinPE, we adopt the parameter values shown to achieve the lower bounds, specifically, $\gamma_1 = 1/M^2, \gamma_2 = 1/(2MK)$. Numerical results for error probability are reported in Section 5. Here, we present the corresponding communication costs for each of the dataset in Figure \ref{app_comm_cost}.

\noindent The numerical results demonstrate that the communication cost of our algorithm \textsc{MaLinBAI-Star} is substantially lower than that of FALinPE \cite{wang2023pure}. Although \textsc{MaLinBAI-Star} incurs a marginally higher communication cost when compared to Async-LinUCB \cite{li2022asynchronous} and FedLinUCB \cite{he2022simple}, it significantly outperforms both in estimating the best arm, highlighting its efficiency in balancing communication overhead with performance in estimating the best arm.

\subsection{Real-World Data}\label{exp:real_world}
In this section, we present the numerical results based on a real-world dataset. For our experiments, we utilize the MovieLens 20M dataset \cite{harper2015movielens}. We follow \cite{wang2023pure} to process the data to fit in the linear bandit setting. Specifically, we retain users with at least 3,000 interactions, resulting in a final dataset comprising 54 users, 26,567 movies, and 214,729 interactions. For each movie, we extract TF-IDF features from the associated tags, producing a feature vector with dimension $d=25$. We then assign a reward $r=1$ to movies with non-zero ratings, and $r=0$ otherwise. Finally, we use these feature vectors and corresponding rewards to learn the true parameter $\theta_*$ via ridge regression.\\

\begin{figure}[!ht]
  \begin{minipage}{0.48\textwidth}
    \centering
    \includegraphics[width=\textwidth]{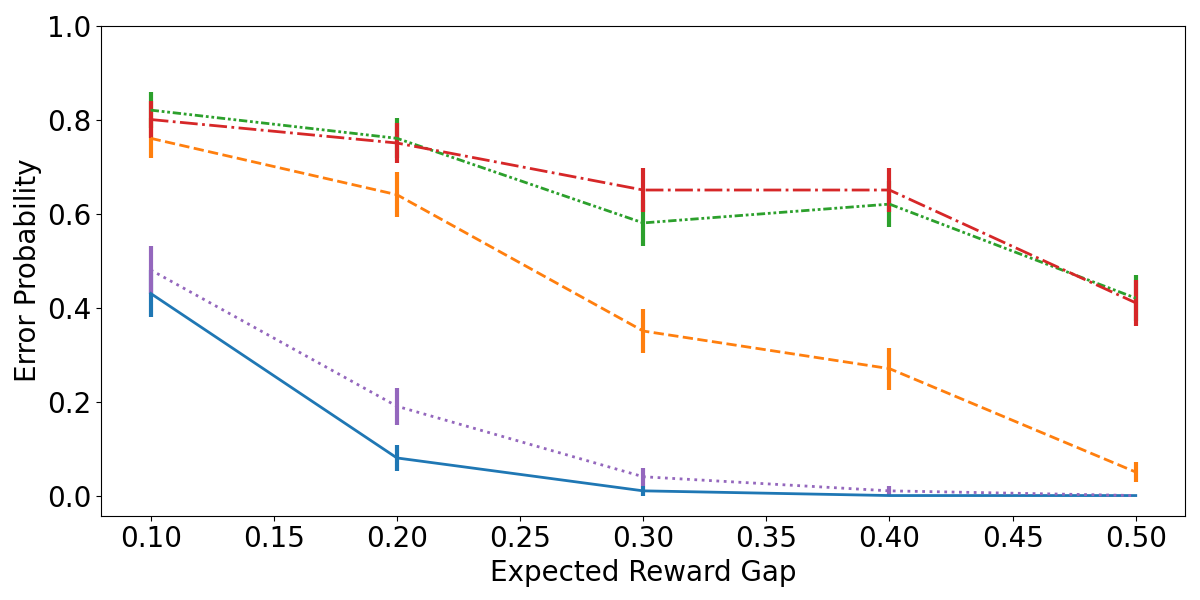}
    \subcaption{Error Probability}
    \label{fig:delta}
  \end{minipage}
  \begin{minipage}{0.48\textwidth}
    \centering
    \includegraphics[width=\textwidth]{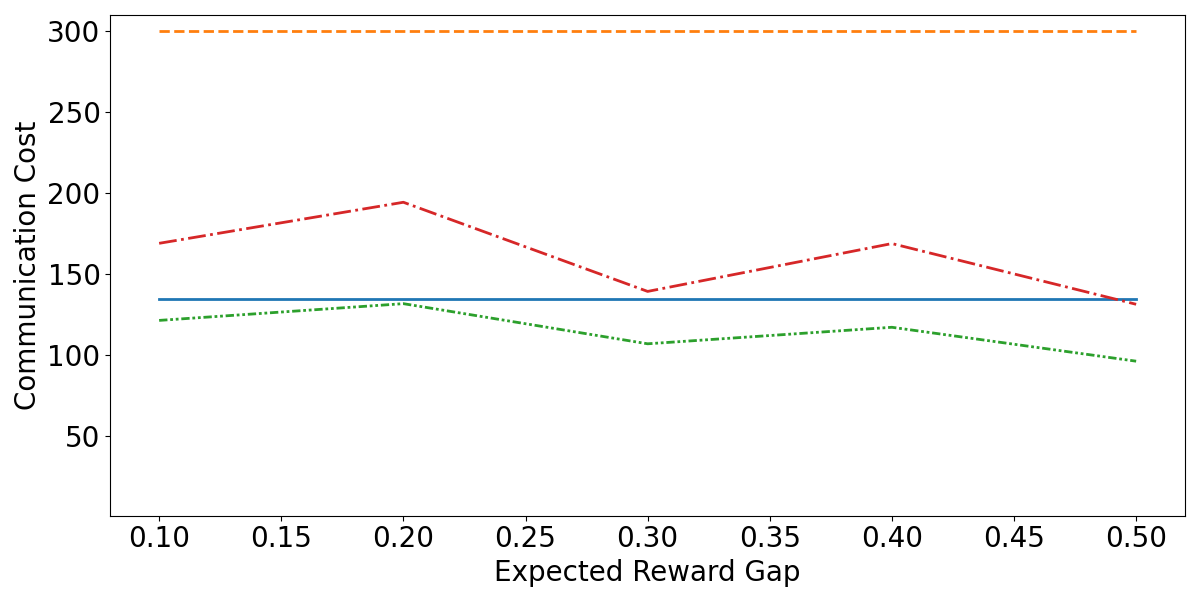}
    \subcaption{Communication cost}
    \label{fig:sphere}
  \end{minipage}
  \centering
    \includegraphics[width=0.8\textwidth]{legend.png}
  \caption{Experimental results for MovieLens data.}
  \label{real_data}
\end{figure}

\noindent We conduct experiments with the expected reward gap varying from $0.1$ to $0.5$. For each reward gap value, we generate a corresponding bandit instance by sampling $k = 10$ vectors from the extracted movie feature vector set, ensuring that the specified reward gap is maintained. All other parameters are kept consistent with the setup detailed in Appendix \ref{exp:comm_cost}. The error probabilities and corresponding communication costs incurred by the algorithms are presented in Figure \ref{real_data}.

\noindent The numerical results show that our algorithm \textsc{MaLinBAI-Star} performs significantly better than FALinPE, Async-LinUCB, FedLinUCB and MA-ODLinBAI in accurately identifying the best arm. Moreover, \textsc{MaLinBAI-Gen} achieves this accuracy with a substantially lower communication cost than FALinPE. Although its communication cost is only slightly higher than that of FedLinUCB, it remains lower than the communication cost of Async-LinUCB. These results demonstrate that our algorithm \textsc{MaLinBAI-Star} consistently outperforms FALinPE, Async-LinUCB, FedLinUCB and MA-ODLinBAI across both real-world and synthetic datasets, all while maintaining efficient communication overhead.\\

\noindent Both of our algorithms \textsc{MaLinBAI-Star} and \textsc{MaLinBAI-Gen} are implemented in Matlab. For the implementation of OD-LinBAI, FALinPE, Async-LinUCB and FedLinUCB we used the code provided in the supplementary material of \cite{yang2022minimax}, \cite{wang2023pure}, \cite{li2022asynchronous} and \cite{he2022simple} respectively. All of our experiments are executed on a computer with 4.05 GHz 8-Core Apple M3 processor and 16 GB memory. 

\end{document}